%% file: nips_cp.tex
\documentclass{article}

\usepackage[final]{nips_2018}

\usepackage{algorithm}
\usepackage{amsfonts}		
\usepackage{amsmath}
\usepackage{booktabs}		
\usepackage{color}
\usepackage{comment}
\usepackage{etoolbox}
\usepackage[T1]{fontenc}	
\usepackage{hyperref}		
\usepackage[utf8]{inputenc}	
\usepackage{microtype}		
\usepackage{nicefrac}		
\usepackage{url}			
\usepackage{wrapfig}

\DeclareMathOperator*{\argmax}{arg~max}

\usepackage{algorithmicx}
\usepackage{algpseudocode}
\usepackage{graphicx}
\usepackage{subcaption}

\newbool{inccomment}
\setbool{inccomment}{true}
\newcommand{\XX}[1]{\ifbool{inccomment}{{\color{magenta} #1}}{}}
\newcommand{\CT}[1]{\ifbool{inccomment}{{\color{magenta}CT\@: #1}}{}}
\newcommand{\NT}[1]{\ifbool{inccomment}{{\color{blue}NT\@: #1}}{}}
\newcommand{\TD}[1]{\ifbool{inccomment}{{\color{orange}#1}}{}}
\newcommand{\FN}[1]{\ifbool{inccomment}{{\color{OliveGreen}#1}}{}}
\newcommand{\GR}[1]{\ifbool{inccomment}{{\color{Tan}#1}}{}}
\newcommand{\LD}{\ifbool{inccomment}{{\color{magenta}\\============================================\\}}}
\newcommand{\RF}{\ifbool{inccomment}{{\color{green}~[R]}}}

\newcommand{\roma}[1]{\uppercase\expandafter{\romannumeral #1\relax}}

\title{Distilling Critical Paths in \\ Convolutional Neural Networks}

\author{
	Fuxun Yu, Zhuwei Qin, Xiang Chen\\
	Department of Electrical and Computer Engineering,
	George Mason University \\
	\texttt{\{fyu2, zqin, xchen26\}@gmu.edu} \\
}

\begin{document}
\maketitle

\input{0_abs}
\input{1_intro}
\input{2_analysis}

\input{3_method}

\input{5_con}

\newpage
\input{ref}

\end{document}

%% file: 0_abs.tex
\begin{abstract}
Neural network compression and acceleration are widely demanded currently due to the resource constraints on most deployment targets.
	In this paper, through analyzing the filter activation, gradients, and visualizing the filters' functionality in convolutional neural networks (CNNs), we show that the filters in higher layers learn extremely task-specific features, which are exclusive for only a small subset of the overall tasks, or even a single class.
	Based on such findings, we reveal the critical paths of information flow for different classes.
	And by their intrinsic property of exclusiveness, we propose a critical path distillation method, which can effectively customize the convolutional neural networks to small ones with much smaller model size and less computation.
\end{abstract}

%% file: 1_intro.tex
\vspace{-3mm}
\section{Introduction}
\label{sec:into}
\vspace{-3mm}

Recently researchers found that the neurons in higher layers of neural networks are usually specialized to encode information relevant to a subset of classification targets [1, 2].
	For example, by ablating certain neurons in higher layers, significant accuracy degradation of specific classes can be observed.
	Based on such intuition, in this paper, we interpret CNN neuron's functionality by activation maximization visualization (AM) [3], and quantifying their contributions to specific classes through activation and gradients analysis.
	With these three analysis approaches, we not only cross-verify the neuron differentiation process across multiple layers, but also reveal that the class relevant information is significantly concentrated in a few neurons.
	In other words, the per-class information flow is directed by a ``critical path'' formed by a few class-specific neurons with significant contributions.


By identifying critical paths, we find that the critical paths for different classes are co-existing but almost exclusive from each other.
	This property of exclusiveness provides the chance to distill the critical paths and further decouple the CNN for individual class decision tasks.
Therefore, we propose our critical path distillation method, which could effectively select out the critical neurons in high layers for a given task, and eliminate the unrelated ones. The distilled neural network contains only critical neurons for single class and perform one-vs-all tasks.
	Compared to conventional compression methods like filter pruning [4], our method proposes a new perspective on per-class information flow and thus could aggressively remove more filters without any retraining.
	Preliminary experiments show that the distilled neural network can have 90\% fewer neurons in higher layers compared to original one, and achieves satisfying performance with 98.6\% average True Positive (TP) rate and 11.0\% False Positive (FP) rate in all one-vs-all tasks.


%% file: 2_analysis.tex
\vspace{-3mm}
\section{Critical Path in Convolutional Neural Networks}
\vspace{-3mm}

\begin{figure}[!tb]
  \vspace{-8mm}
  \centering
  \includegraphics[width=5.5in]{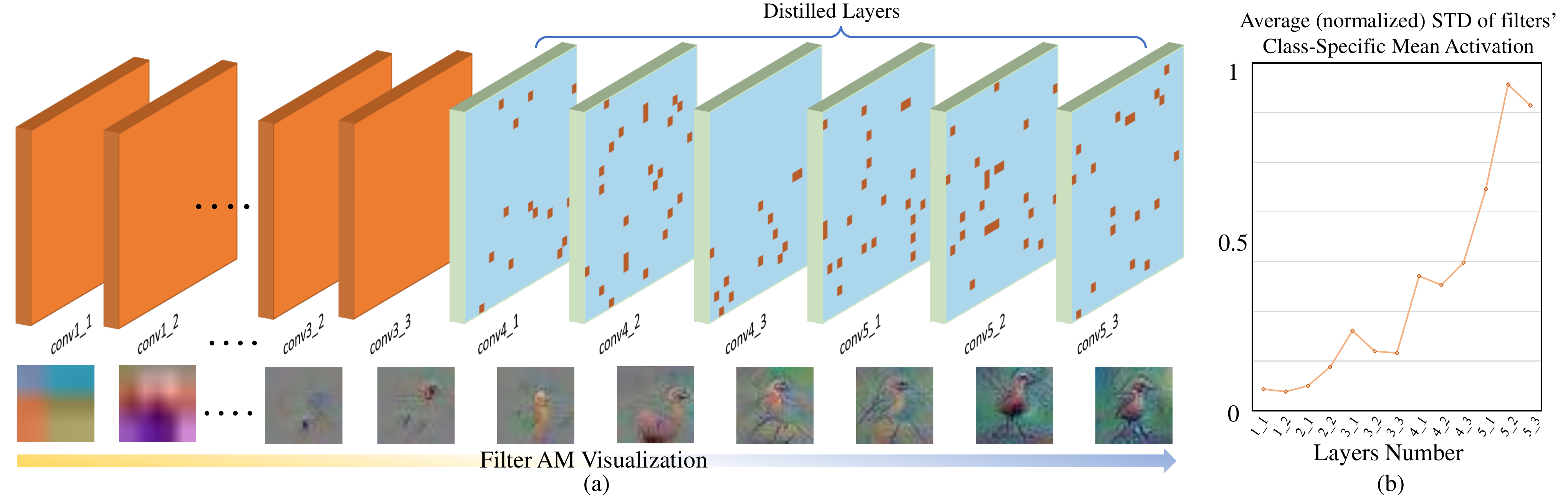}
  \vspace{-5mm}
  \caption{(a) Critical Path Overview. Filters begin to differentiate and form more specific functions in the latter layers. (b) Average (normalized) standard deviation (STD) of filters’ per-class mean activations, which also show that high layer filters have strongly biased class preference.
}
  \vspace{-3mm}
  \label{fig:cp}
\end{figure}

\textbf{2.1 Overview of Critical Path}
Fig.~\ref{fig:cp} (a) shows a conceptual critical path for the ``bird'' class (CIFAR10/VGG-16) [5].
    The orange color in the layers denote the reserved neurons from the critical path, while all the blues ones are removed from the original network due to their insignificance to the ``bird'' class. Our critical path contains two clear main components: full basic feature layers (low) and distilled class-specific neurons in high layers.
  In lower layers, all filters are reserved since they usually learn basic and class-agnostic features, e.g. colors, edges [6, 7]. 
  However, in higher layers, filters are gradually differentiated to recognize different classes, as shown by their AM patterns. Therefore, only the class-related neurons are reserved in the critical path. Fig.~\ref{fig:cp} (b) also verifies this by showing the trend of standard deviation (STD) of filters' per-class mean activation from low to high layers. High layer filters averagely have the larger standard deviation, which means their activation for different class inputs is highly non-uniform. 
  In our experiments, we find that most neurons in high layers are mostly activated by a single class, which therefore forms very exclusive critical paths for different classes. Furthermore, the critical path for a single class can be decoupled from the original neural network and acts as a much smaller but task-specific neural network, which we will show in the later section.

\textbf{2.2 Critical Path Interpretation and Verification through Neuron Contribution Analysis}
The identification of critical path is based on the neuron contribution significance to specific classes.
	In this work, we include three approaches to analyze the neurons' functionality, and show that the cross-verification of the three approaches testifies the proposed critical path.

Suppose a CNN's logit layer output could be formulated as $Z(x)$ with $n$ logits, which is corresponding to $n$ classes. Given a specific layer $l$ with $I_{l}$ filters, the $n_{th}$ logit output $Z_n$ is:
\begin{equation}
\small
	Z_n(W, A^{l}) = W^L (... ~\alpha(W^{l+1} * A^{l}+b^{l+1}) ...) + b^L,
	\label{eq:1}
\end{equation}
\normalsize
where $W$ and $b$ are layer weights and biases. Then we can evaluate neurons' contribution by:

\textbf{(1) Mean Absolute Activation} For a filter $F_i$ in layer $l$, its activation can be denoted by $A^{l}_i$. The first method we use is the mean absolute activation $A^{l}_i$ with regard to different classes inputs. When $W^{l+1}_{i}$ is non-zero, larger $A^{l}_i$ means more contribution to the final $Z_n$, as shown in Eq.~\ref{eq:1}. Therefore, the class $C$ which produces the largest mean absolute activation could be considered as its most important functionality or task since these images could mostly activate the filter.

\textbf{(2) Contribution Index} Meanwhile, we could also use first-order Taylor Approximation [8] to approximate $Z_n$ with regard to filter $F_i$'s activation $A^l_i$:
\begin{equation}
\small
\vspace{-3mm}
	Z_n(A^{l}_i + \delta) = Z_n(A^l) + \frac{\partial Z_n(A^l_i)}{\partial A^l_i} \cdot \delta.
	\label{eq:2}
\end{equation}
\normalsize
\vspace{-2mm}

Here, we use the $\ell_1$ of the coefficient matrix ${\partial Z_n(A^l_i)}/{\partial A^l_i}$ as the contribution index of $i_{th}$ filter's output to the specific class $n$: Larger value means a small $\delta$ change of $A^{l}_i$ will be amplified to cause larger influence of $Z_n$. One filter has different contribution indexes to different classes. Therefore, the class $C$ with largest contribution index could be also considered as its most important function.

\textbf{(3) AM Visualization} Lastly, we introduce the Activation Maximization (AM) Visualization method [3]. AM is a well-established method to interpret a single filter's activation preference. It adopts gradient ascent algorithm to maximize a neuron's activation with input $X$:
\begin{equation}
\small
	P =\argmax_{X} {A^l_i(X),
	\hspace{0.6cm} X_{k+1} \leftarrow X_{k}} + \eta \cdot \frac{\partial A^l_i(X_{k})}{ \partial X_{k}},
	\label{eq:am}
\end{equation}
\normalsize
where $\eta$ is the step size, and the final $P$ is the visualized pattern of filter $i$ in layer $l$. The crafted pattern $P$ could also maximize its activation and thus could visually interpret the filter's functionality.

\begin{figure}[!tb]
  \vspace{-8mm}
  \centering
  \includegraphics[width=5.5in]{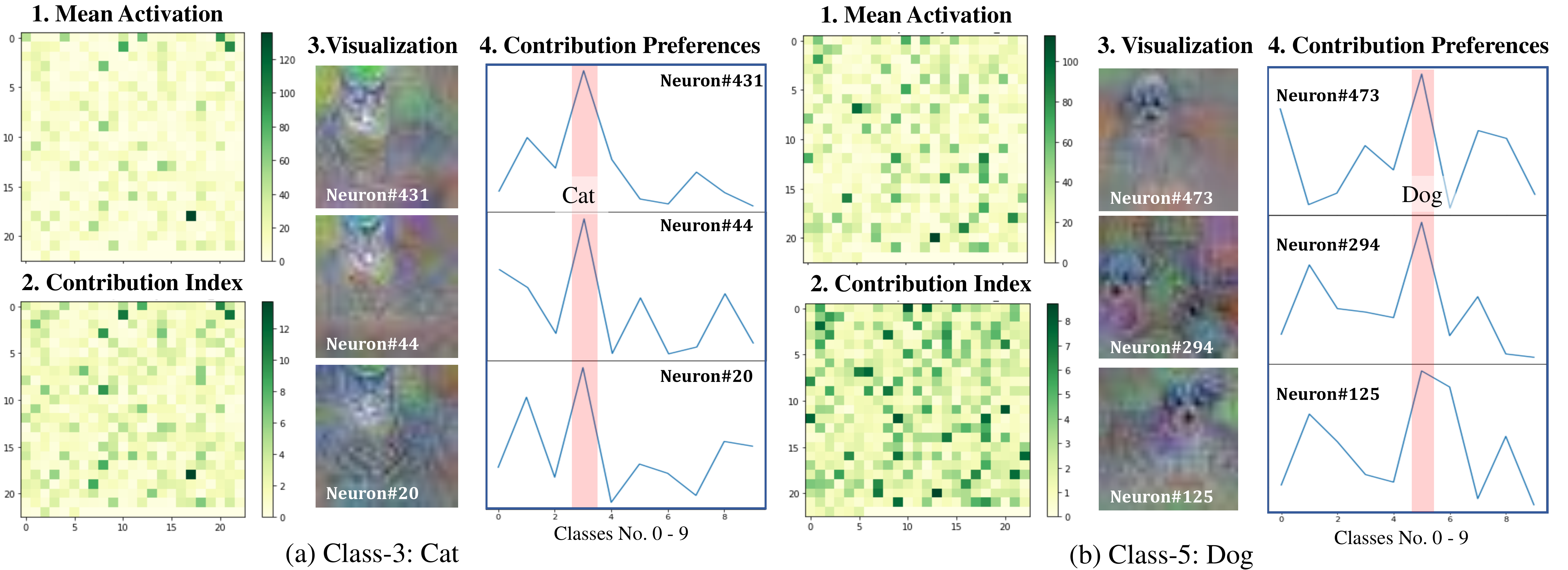}
  \vspace{-5mm}
  \caption{Mean activation map, contribution index map and AM filter visualization for filters in the Conv5\_1 layer of VGG16. The 3rd and 5th classes in CIFAR10 are selected for evaluating. Three evaluation metrics cross-verify the existence of critical paths for the single specific class. }
  \vspace{-5mm}
  \label{fig:act_grad_vis}
\end{figure}

\textbf{2.3 Critical Path Verification}
Next, we combine the three different metrics together to cross-verify the filter functionality analysis and demonstrate the existence of the critical path for the single class. The convolutional neural network we used here is VGG16 trained on CIFAR10 same as [4].  We choose the $10_{th}$ layer Conv5\_1 and class $3$ and $5$ (cat and dog) as the example (Other classes also share the same properties), and results are shown in Fig.~\ref{fig:act_grad_vis} (a) and (b).

From the mean absolute activation map and contribution index map of Fig.~\ref{fig:act_grad_vis} (a), we could clearly see that they point out similar filter groups with deeper colors: The filters with the largest activation also have the maximal contribution indexes among all the filters. According to Taylor expansion in Eq.~\ref{eq:2}, both larger activation and coefficients will lead to more contribution to the final decision of this class. Thus we call these filters the critical neurons, which form the critical path of this class. Here we also visualize the top three filters with the largest product of mean activation and contribution index, which show exactly the cat patterns. We also show the three filters' contribution indexes to all classes. The biggest contribution indexes of three filters all lie in the cat class and are much larger than other classes. For generality, we also show the dog class's critical path analysis in Fig.~\ref{fig:act_grad_vis} (b), where all similar conclusions could be drawn. Next, we will introduce our critical path distillation method.


%% file: 3_method.tex
\vspace{-3mm}
\section{Critical Path Distillation}
\label{sec:distillation}
\vspace{-3mm}

In this section, we will introduce how to distill the critical path of a single class. First, we need to choose layers in which filters have formed specific class functionality. The higher layers we choose, the critical paths distillation could be easier but the model size compression effect may be lower. According to the experiments, we empirically choose the last six convolutional layers for distillation.
Given a specific class, we first compute the corresponding mean activation map and contribution index map of all six layers. Maps of their product are also calculated to combine these two factors to choose reserved filters. The reserved ratio controls the ``width'' of the critical path, and trade-off the accuracy and compression rate. For example, $10\%$ reserved ratio means we only keep 52 filters our of 512 in all the chosen layers. A ratio below can cause catastrophic accuracy drop.
Another important thing is since we modify the neural network structure, the previous ten biases of logit layer are not suitable. Therefore, we set the biases in logit layer to zeros. Without any retraining, the distilled model preserves the ability well to respond to the chosen class and reject all other classes, i.e. one-vs-all tasks. We use TP (True Positive) rate and FP (False Positive) rate to evaluate our critical path distillation performance on ten one-vs-all tasks, respectively. The final results are shown in Table.1. And the final model size is compressed to 11.4 MB compared to original 59.0 MB. 

\begin{table}[!tb]
\centering
\caption{Results for Single Class Critical Path Distillation (Reserved Ratio: 10\%)}
\begin{tabular}{|l|l|l|l|l|l|l|l|l|l|l|}
\hline
Class & 0    & 1    & 2    & 3    & 4    & 5    & 6    & 7    & 8    & 9    \\ \hline
TP    & 1.0  & 1.0  & 0.99  & 0.95  & 1.0 & 0.98  & 0.981 & 1.0 & 0.98  & 0.98  \\ \hline
FP    & 0.114 & 0.126 & 0.109 & 0.096 & 0.110 & 0.091 & 0.118 & 0.114 & 0.108 & 0.113 \\ \hline
\end{tabular}
\vspace{-5mm}
\end{table}

%% file: 5_con.tex
\vspace{-3mm}
\section{Conclusion}
\vspace{-3mm}

In this work, we propose a critical path distillation method in convolutional neural networks. Through analyzing the activation, gradients and visualization patterns, we demonstrate that higher layer filters always learn class-specific features and therefore could be well distilled to customize class-specific neural networks. Our preliminary experiments show that the customized class-specific neural networks by critical path distillation could achieve average 98.6\% TP rate and 11.0\% FP rate in one-vs-all tasks, as well as have much smaller size and computation requirements. In the future, we will generalize our method to multiple classes and on the large-scale dataset, ImageNet [9].

%% file: ref.tex
\section*{References}
\small

[1] Zhou, B., Sun, Y., Bau, D. and Torralba, A., 2018. Revisiting the Importance of Individual Units in CNNs via Ablation. arXiv preprint arXiv:1806.02891.

[2] A. S. Morcos, D. G. Barrett, N. C. Rabinowitz, and M. Botvinick. On the importance of single directions for generalization. International Conference on Learning Representations, 2018.

[3] Yosinski, Jason, et al. "Understanding neural networks through deep visualization." arXiv preprint arXiv:1506.06579 (2015).

[4] Li, Hao, et al. "Pruning filters for efficient convnets." arXiv preprint arXiv:1608.08710 (2016).

[5] Simonyan, Karen, and Andrew Zisserman. "Very deep convolutional networks for large-scale image recognition." arXiv preprint arXiv:1409.1556 (2014).

[6] Krizhevsky, Alex, Ilya Sutskever, and Geoffrey E. Hinton. "Imagenet classification with deep convolutional neural networks." Advances in neural information processing systems. 2012.

[7] Yosinski, Jason, et al. "How transferable are features in deep neural networks?." Advances in neural information processing systems. 2014.

[8] LeCun, Yann, John S. Denker, and Sara A. Solla. "Optimal brain damage." Advances in neural information processing systems. 1990.

[9] Deng, Jia, et al. "Imagenet: A large-scale hierarchical image database." Computer Vision and Pattern Recognition, 2009. CVPR 2009. IEEE Conference on. Ieee, 2009.